\documentclass[a4paper,conference]{IEEEtran}
\usepackage{cite}

\ifCLASSINFOpdf

  \usepackage[pdftex]{graphicx}
  \graphicspath{ {images/} }
\else
\fi
\usepackage{amsmath}
\usepackage{amssymb}
\usepackage{tablefootnote}
\usepackage{tikz}
\usetikzlibrary{backgrounds}
\usepackage{subcaption,mwe}

\makeatletter
\DeclareRobustCommand\onedot{\futurelet\@let@token\@onedot}
\def\@onedot{\ifx\@let@token.\else.\null\fi\xspace}

\def\eg{\emph{e.g}\onedot} 
\def\ie{\emph{i.e}\onedot} 
 
\def\etc{\emph{etc}\onedot} 
 
\def\etal{\emph{et al}\onedot}
\makeatother

\def\todo[#1]{\textcolor{red}{TODO: #1}}

\begin{document}
%
\title{3D Human Pose Estimation from \\ Deep Multi-View 2D Pose}

\author{\IEEEauthorblockN{Steven Schwarcz}
\IEEEauthorblockA{
University of Maryland\\
College Park, Maryland 20742\\
Email: schwarcz@umiacs.umd.edu}
\and
\IEEEauthorblockN{Thomas Pollard}
\IEEEauthorblockA{Systems and Technology Research (STR)\\
Woburn, Massachusetts\\
Email: tom.pollard@stresearch.com}}


%


\maketitle

\begin{abstract}
Human pose estimation - the process of recognizing a human's limb positions and orientations in a video - has many important applications including surveillance, diagnosis of movement disorders, and computer animation.  While deep learning has lead to great advances in 2D and 3D pose estimation from single video sources, the problem of estimating 3D human pose from multiple video sensors with overlapping fields of view has received less attention.   When the application allows use of multiple cameras, 3D human pose estimates may be greatly improved through fusion of multi-view pose estimates and observation of limbs that are fully or partially occluded in some views.  Past approaches to multi-view 3D pose estimation have used probabilistic graphical models  to reason over constraints, including per-image pose estimates, temporal smoothness, and limb length.  In this paper, we present a pipeline for multi-view 3D pose estimation of multiple individuals which combines a state-of-art 2D pose detector with a factor graph of 3D limb constraints optimized with belief propagation.  We evaluate our results on the TUM-Campus and Shelf datasets for multi-person 3D pose estimation and show that our system significantly out-performs the previous state-of-the-art with a simpler model of limb dependency.
\end{abstract}


%
\IEEEpeerreviewmaketitle








\section{Introduction}

Articulated human pose estimation is a long-studied problem in the field of computer vision \cite{Felzenszwalb:2005:PSO:1024426.1024429,5206754} which involves estimating a parameterized 2D or 3D human body model from video or still imagery.  Pose estimation has many direct applications including motion capture for film and game production and may also be done as an pre-precessing step for higher-level vision problems such as action recognition.  Most of the recent progress made in this area has leveraged advances in deep neural networks, particularly convolutional networks \cite{Newell2016}, to achieve impressive results in the areas of 2D and 3D pose estimation from single images \cite{Newell2016,Cao2016}.

While significant breakthroughs, the utility of these approaches are limited by the single-view nature of the source data.  2D pose estimation techniques only detect joint and limb locations in image space, and cannot reveal anything directly about the positions and orientations of limbs in 3D.  Approaches exist which perform 3D pose estimation on a single actor in a single image \cite{DBLP:journals/corr/PavlakosZDD16}, but the range of 3D poses is heavily limited by the training dataset and limbs which are occluded must be inferred rather than directly measured.  For applications which demand precise measurement of all 3D limbs, a synchronized network of overlapping calibrated cameras can be used to resolve single-view ambiguities and improve 3D triangulation accuracy.   


To accelerate research in multi-person, multi-camera 3D pose estimation, Amin, Andriluka, Rohrbach, Schiele introduced the TUM Campus and Shelf datasets \cite{Amin2013,Belagiannis16}, each of which contain video footage of several actors taken from 2-3 calibrated cameras.  This dataset presents an opportunity to face challenges not present in single image pose estimation.  For example, actors in the scene are frequently fully occluded by other actors in one view, and in this situation it becomes necessary to keep track of the actors through the occlusion event.

To address these challenges \cite{Amin2013}, and later \cite{Belagiannis16}, use the sum-product belief propagation algorithm to optimize over the space of 3D pose configurations, which they discretize by triangulating the results of a pose detector.  The factor graph they use features a large number of parameters for factors such as joint collision and temporal smoothing. For their system to achieve optimal results, however, they need to use a structured SVM solver to optimize over a series of hyper parameters \cite{Belagiannis16}.

The system we propose, by contrast, achieves significantly better results with fewer constraints on limb motion.  First, we use a modern 2D pose detector \cite{Cao2016} which leverages advances in deep learning to generate better 2D limb hypotheses.  Next we use a factor graph optimization, where factors are all constructed to be readily interpretable, and as such it is very easy to identify reasonable values for each factor without the need to resort to a more complex hyperparameter optimization scheme.  Our final factor graph uses only three factors: limb location priors, collision terms, and temporal smoothing.   

In this paper, we present our approach: a full pipeline for multiple person, multiple camera 3D pose estimation.  After presenting our technique, we empirically show the effectiveness of our technique on the Campus and Shelf datasets, where we in some cases significantly out-perform the previous state-of-art.

\section{Related Work}

\subsection{2D Human Pose Estimation}

Approaches to human pose estimation can be categorized broadly into two separate tasks based on goals: single person pose estimation and multiple person pose estimation.  In single person pose detection, a tight bounding box is provided around the person in question.  Recently, stacked hourglass networks \cite{Newell2016} and their variants have emerged as the leading deep learning methods for single person pose estimation.  A stacked hourglass network is a fully-convolutional CNN architecture with multiple "hourglass modules" which squeeze the image representation to a very small size, allowing for each unit in the CNN to have a very large receptive field.






Alternatively, multi-person pose estimation seeks a single architecture which, given a single image with multiple people present, can identify the individual people before or during the actual pose estimation process.  These techniques can be "top-down", where people are first identified and then their poses are detected on an individual basis \cite{DBLP:journals/corr/PapandreouZKTTB17}, or "bottom-up", where joints or limbs are detected first, and then aggregated into human detections only afterwards.  To this end, Cao, Simon Wei, and Sheikh present Part Affinity Fields \cite{Cao2016}, which, in addition to detecting individual joints, also detects a vector indicating the orientation of each joint.  These joints are then aggregated into separate skeletons, allowing for simultaneous person and part detection.


\subsection{3D Human Pose Estimation}

Exploiting the success of 2D human pose estimation techniques, there has been a lot of work done in the area of 3D pose estimation, as inferred from a single image \cite{Zhou2017}.  For example, \cite{DBLP:journals/corr/PavlakosZDD16} uses a modified stacked hourglass to progressively produce 3D pose detections with more and more depth. 

3D pose estimation results can be improved when multiple cameras are available. \cite{Sigal:2012:LPE:2205801.2205829} use a graphical model for 3D pose estimation and perform inference with a particle message passing scheme. \cite{7299005} use a Sum of Gaussian based appearance model, which is later expanded in \cite{Elhayek:2017:MMM:3058313.3058397} to make use of deep learned features.

There also exist multiple datasets for 3D human pose estimation in controlled indoor environments \cite{Sigal:2006:MLR:1153171.1153701,Sigal2009, h36m_pami}, where multiple cameras observe a single actor in a studio.  Many alternate datasets address the problem of 3D human pose detection with multiple humans, or in outdoor environments \cite{Amin2013}.  In particular, \cite{Amin2013} introduces the TUM Campus and Shelf datasets, each of which feature three actors in a single video taken from at least three synchronized cameras.  Following up on this, \cite{Belagiannis14} and \cite{Belagiannis16} use factor graphs optimized with belief propagation to perform inference on poses detected with 2D pose estimators.


\section{Approach}

The human body is represented as a graph with $N = 14$ nodes, where each node represents a joint in the human body (\eg left hand, right shoulder, top of head, \etc), and edges represent adjacent joints.  Our objective is to estimate, for each individual and timestep $t=0...T$, the 3D joint locations $\mathbf{Y}^t = \{y^t_i|i=1, 2, \ldots, N\}$ where each $y^t_i \in \mathbb{R}^3$. The input data comes in the form of video frames $\{I_c^t\}$ from each camera $c$, which have been pre-calibrated such that projection matrices have been computed and video frames have been aligned temporally.  

Our processing pipeline for estimating the $\{\mathbf{Y}^t\}$ poses for each person from the video has multiple stages.  First, 2D detections of joints are computed for each individual in the image using a state-of-art detector.  When multiple people are present in the scene, these detections are used to identify specific individuals by triangulating 2D joint detections from different views.  For each individual, a conditional random field (CRF) is constructed over 3D limb location variables to model the dependency on information sources including 2D detection strength, temporal smoothness, and collision terms. Inference on each factor graph is performed using the sum-product belief propagation algorithm to obtain posterior probabilities on limb location, from which the maximum likelihood 3D location is selected. These steps are described in detail in the following subsections.

\subsection{2D Pose Detection and Cross-frame Association}

\begin{figure}
\centering
\begin{subfigure}[t]{.16\textwidth}
\includegraphics[scale=0.7]{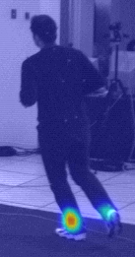}
\caption{Heatmap for 2D detection}
\label{fig:heatmap}
\end{subfigure}
\qquad
\begin{subfigure}[t]{.16\textwidth}
\includegraphics[scale=0.7]{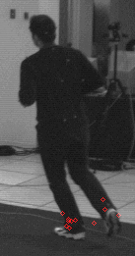}
\caption{Sampling heatmap}
\label{fig:sampling}
\end{subfigure}
\caption{(a) OpenPose outputs a heat map $H^t_{c, i}$ for each image, where intensity corresponds to the likelihood that a particular joint exists in that location.  
(b) In order to generate a set of possible states in $\mathbb{R}^3$ for some joint $y^t_i$, we convert  $H^t_{c, i}$ into a probability mass function and sample from it.}
\label{fig:globfig}
\end{figure}

The first step in our pipeline is to obtain 2D limb detections in each frame of video from each camera.  Many published approaches to this problem have made their software available and we choose to use the open source OpenPose library \cite{Cao2016}, which uses part-affinity fields to simultaneously extract human poses and detect individuals.  Along with offering state-of-art 2D detection performance, this software can expose raw limb likelihood heat maps (see Figure \ref{fig:heatmap}) used to compute the final 2D limb position.  These heat maps  $\mathbf{H}^t_c = \{H^t_{c, 1}, H^t_{c, 2}, \ldots, H^t_{c, N}\}$ allow propagation of uncertainty in the 2D detection step to the 3D optimization step, where a final decision can be made with all multi-view evidence.  Each of these heat maps $H^t_{c, i}$ assigns a value in $[0, 1]$ to each pixel in $I^t_c$ indicating the likelihood that the given pixel is displaying the projection of joint $i$. 
%
%
%
%

Given a collection of 2D poses within each frame, we must match skeletons belonging to the same person both across time and multiple views.  Consider an individual $G^t_c = \{g^t_{c, 1}, g^t_{c, 2}, \ldots g^t_{c, N}\}$ detected in 2D at time $t$ for camera $c$. Here, each $g^t_{c, i}$ represent the pixel locations of a 2D joint, \ie a local maximum of the corresponding $H^t_{c, i}$.  We determine that two detected individuals $G^t_c$ and $G^{t+1}_c$ from two consecutive frames are actually the same person if their bounding boxes have an intersection over union (IoU) score of at least $0.7$.   If a detection $G^t_c$ does not have an IoU match with any bounding box in the previous frame, then we determine it to be the first occurrence of that individual.

To determine matches between different cameras at the same point in time, \ie between two poses $G_c^t$ and $G_{c'}^t$, we cast a ray into the scene for each pair of joints $g^t_{c, i}$ and  $g^t_{c', i}$ in the pose, and measure the distance between the closest points of each ray (\eg, the distance between the wrist detection in  $G_c^t$ and the same detection in $G_{c'}^t$).  If the average distance between these points is less than 2cm for all joints that are visible in both skeletons, we determine that the two skeletons belong to the same person.  

When we have completed the matchings, we remove from each frame any identities that only appear in a single camera, leaving behind only detections that can be fed directly into our multi-view system.

\subsection{Conditional Random Field for 3D Pose}
At this stage, the joint locations from matched detections in multiple views can be back-projected into 3D and triangulated to obtain a 3D pose estimate.  The result is noisy, lacks temporal smoothness frame-to-frame, and may occasionally contain gross errors for certain limbs if 2D detections are ambiguous.  To improve the quality of the pose estimates, a CRF on 3D joint positions is constructed to allow the introduction of additional consistency constraints. The CRF is represented by a factor graph, where the joint positions are the variables and constraints on these joints become factors:

\begin{multline} \label{eq:full}
p(\{\mathbf{Y}^t\}|\{I_c^t\}) = 
\\\frac{1}{Z} \prod^T_{t=1} \prod^N_{i=1} f_\mathit{data} (y^t_i,\{I_c^t\}) \cdot f_\mathit{temp} (y^t_i, y^{t-1}_i, y^{t+1}_i) \cdot
\\  \prod^T_{t=1} \prod_{(i, j) \in E_{col}} f_\mathit{col} (y^t_i, y^t_j)
\end{multline}

\noindent where $Z$ is the partition function. The factors $f$ are not necessarily true probability distributions although we do individually normalize all of our factors into probabilities, as we found that doing so leads to improved numerical stability.

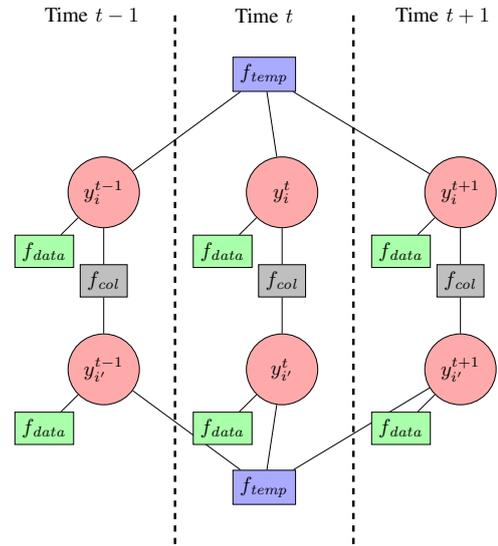
\begin{figure}

\centering
\resizebox{!}{.4\textwidth}{
    \begin{tikzpicture}
    \node[] at (-.2,9) {Time $t-1$};
    \node[] at (2.7,9) {Time $t$};
    \node[] at (5.7,9) {Time $t+1$};
    \path 
    (2.7, 8) node[fill={rgb:blue,1;white,2}, rectangle,draw](t1) {$f_\mathit{temp}$}
    (0,6) node[fill={rgb:red,1;white,2}, circle,draw, minimum size=1.2cm](y1) {$y^{t-1}_i$}
    (3,6) node[fill={rgb:red,1;white,2}, circle,draw, minimum size=1.2cm](y2) {$y^{t}_i$}
    (6,6) node[fill={rgb:red,1;white,2}, circle,draw, minimum size=1.2cm](y3) {$y^{t+1}_i$}
    (0,4.5) node[fill={rgb:black,1;white,3}, rectangle,draw](c1) {$f_\mathit{col}$}
    (3,4.5) node[fill={rgb:black,1;white,3}, rectangle,draw](c2) {$f_\mathit{col}$}
    (6,4.5) node[fill={rgb:black,1;white,3}, rectangle,draw](c3) {$f_\mathit{col}$}
    (0,3) node[fill={rgb:red,1;white,2}, circle,draw, minimum size=1.2cm](y4) {$y^{t-1}_{i'}$}
    (3,3) node[fill={rgb:red,1;white,2}, circle,draw, minimum size=1.2cm](y5) {$y^{t}_{i'}$}
    (6,3) node[fill={rgb:red,1;white,2}, circle,draw, minimum size=1.2cm](y6) {$y^{t+1}_{i'}$}
    (2.7, 1) node[fill={rgb:blue,1;white,2}, rectangle,draw](t2) {$f_\mathit{temp}$}
    (-1,5) node[fill={rgb:green,1;white,2}, rectangle,draw](d1) {$f_\mathit{data}$}
    (2,5) node[fill={rgb:green,1;white,2}, rectangle,draw](d2) {$f_\mathit{data}$}
    (5,5) node[fill={rgb:green,1;white,2}, rectangle,draw](d3) {$f_\mathit{data}$}
    (-1,2) node[fill={rgb:green,1;white,2}, rectangle,draw](d4) {$f_\mathit{data}$}
    (2,2) node[fill={rgb:green,1;white,2}, rectangle,draw](d5) {$f_\mathit{data}$}
    (5,2) node[fill={rgb:green,1;white,2}, rectangle,draw](d6) {$f_\mathit{data}$}
    ;
    \draw[dashed, very thick](1.2, 9) -- (1.2, 0);
    \draw[dashed, very thick](4.2, 9) -- (4.2, 0);
    \begin{pgfonlayer}{background}
    \draw (t1) -- (y1);
    \draw (t1) -- (y2);
    \draw (t1) -- (y3);
    \draw (t2) -- (y4);
    \draw (t2) -- (y5);
    \draw (t2) -- (y6);

    \draw (c1) -- (y1);
    \draw (c2) -- (y2);
    \draw (c3) -- (y3);
    \draw (c1) -- (y4);
    \draw (c2) -- (y5);
    \draw (c3) -- (y6);

    \draw (d1) -- (y1);
    \draw (d2) -- (y2);
    \draw (d3) -- (y3);
    \draw (d4) -- (y4);
    \draw (d5) -- (y5);
    \draw (d6) -- (y6);
    \end{pgfonlayer}
    \end{tikzpicture}
    }
   
   
    
    
     

   \caption{A factor graph diagram for a simplified network involving two symmetric joints $y^t_i$ and $y^t_{i'}$ and three timesteps.  The joint variables are denoted with circles and the data, temporal, and collision factors are denoted with green, purple, and gray rectangles, respectively.}
   

\label{fig:factGraph}
\end{figure}

The first of these factors, $f_\mathit{data}(y_i^t)$, represents the likelihood of a joint $y^t_i$ given evidence aggregated from the detections in the source images $\{I_c^t\}$.  The second factor is $f_\mathit{temp}$, which encourages the movement of joints between time steps to be temporally smooth. $f_\mathit{col}$ constrains certain symmetric pairs of joints not to collide.  The set of edges $E_\mathit{col}$ represents the joints that cannot collide, which in our implementation represent symmetric body parts, \eg right and left wrist, right and left ankle, \etc.  An illustration of the dependencies of these factors on variables in Equation \ref{eq:full} is shown in Figure \ref{fig:factGraph} for a subset of the full body network.  Additional constraints, such as constant limb lengths, may be easily appended to the factorization if desired.  

To make inference on the network tractable, the joint positions $\{y^t_i\}$ are limited to take on a discrete number of possible states.   To discretize the state space for $\mathbf{Y}$, we use a method similar to that of\cite{Belagiannis16}.   Let $\mathbf{S}_i^t = \{s_{i, 1}^t, s_{i, 2}^t, \ldots, s_{i, M}^t\}$ be the discrete set of $M$ possible states that $y_i^t$ can take on. Our goal is to a select a small set of $s_{i, j}^t \in \mathbb{R}^3$ that well sample the 2D pose heat maps $\{H^t_{c, i}\}$ when projected into each image.  

We iteratively select a subset of cameras of size $C'$ with $2 \leq C' \leq C$ (a given joint $i$ will generally be visible from either 2 or 3 cameras in the data we have). We then randomly sample from the 2D heat map $H^t_{c, i}$ in each camera and back-project the pixel coordinate into a ray.  A 3D joint hypothesis $s_{i, j}^t$ is formed by computing the point in $\mathbb{R}^3$ closest to all $C'$ rays.  For our implementation, we draw 16 samples this way when $C = 3$ and 64 samples when $C = 2$.  In both cases, the final result is $M=64$ samples, because when $C=3$, there are three subsets of cameras with $C' = 2$ and one with $C'=3$.  See Figure\ref{fig:sampling}.  

In the subsections that follow, we provide detailed descriptions of the potential functions in (\ref{eq:full}) and our CRF optimization procedure using the sum-product belief propagation algorithm \cite{910572} to obtain posterior estimates on the joint positions. 

\subsubsection{Data Term}

The data term $f_\mathit{data} (y^t_i)$ in the factor graph measures 2D pose confidence of the projected 3D joint hypothesis.  This term is straight-forward to calculate: for each possible state $s_{i, j}^t$, we calculate $f_\mathit{data}(s_{i, j}^t)$ by projecting $s_{i, j}^t$ from world coordinates onto each $H^t_{c, i}$ and averaging the results:

\begin{align}
f_\mathit{data}(s_{i, j}^t) = \frac{1}{C} \sum_{c=1}^C H_{c, i}^t(\hat{s}_{i, j}^t)
\end{align}

\noindent where $\hat{s}$ is the projection of $s$ in the image.  Referring back to the 2D pose heat maps, as opposed to the single optimal 2D locations, allows this data term to fuse weak evidence in the case of ambiguity, such as bimodal distributions around left/right joints in close proximity to each other.

\subsubsection{Temporal Term}

The purpose of the temporal term $f_\mathit{temp}$ is to bias joints to move smoothly frame-to-frame.  Without some form of temporal smoothing, we found that animations of the estimated 3D body models contained significant jitter as a result of each frame being calculated separately.  Having information flow between time steps also helps the system to recover from short-term confusion events such as occlusion of limbs by the torso during walking.

We model temporal smoothing with a ternary term, involving 3D joint estimates from three consecutive time steps: $y^t_i, y^{t-1}_i$, and $y^{t+1}_i$.  Specifically: 


\begin{multline}
f_\mathit{temp} (s_{i, j}^t, s_{i, j'}^{t-1}, s_{i, j''}^{t+1}) = 
\\ \exp{\left[-\frac{1}{2\sigma_\mathit{temp}^2} \lVert s_{i, j}^t - \mu(s_{i, j'}^{t-1},s_{i, j''}^{t+1}) \rVert\right]}
\end{multline}

\noindent where $\mu(s_{i, j'}^{t-1},s_{i, j''}^{t+1}) = \frac{s_{i, j'}^{t-1} + s_{i, j''}^{t+1}}{2}$ is the expected position of the joint using a constant velocity assumption and samples immediately before and immediately after the current time step.  A state hypothesis $s^t_{i, j}$ is penalized if it veers too far from the constant-velocity prediction.  We experimentally set $\sigma_\mathit{temp}=2\text{cm}$, though we found that the results were not very sensitive to the exact choice of $\sigma_\mathit{temp}$ so long as the choice was reasonably small.  



\subsubsection{Collision Term}

The collision term $f_\mathit{col}$ prevents symmetric left/right joints from colliding with each other in 3D.  This common issue in pose detection is generally caused by the fact that symmetric joints, like the right and left feet, look very similar to each other and are often in close proximity.  This tends to result in a bimodal distribution in the 2D pose heat maps, which in turn can cause both symmetric joints in a pair to be assigned to the same location in 3D space.

In contrast to\cite{Belagiannis16}, who handle collision with a trained Gaussian on the distance between limbs, we instead use a modified sigmoid function that evaluates to 1 when the two points are reasonably far apart, and 0 when they are too close:

\begin{align}
f_\mathit{col} (s_{i, j}^t, s_{k, j}^t) = \frac{1}{1 + e^{\theta_1-\theta_2\lVert s_{i, j}^t - s_{k, j}^t \rVert}}
\end{align}

\noindent where $\theta_1$ is a hyperparameter that offsets the center of the sigmoid, and $\theta_2$ is a hyperparameter that tightens the sigmoid.  We empirically set $\theta_1 = 15\text{cm}$ and $\theta_2 = 10\text{cm}$.  As with $\sigma_\mathit{temp}$, we find that as long as the values chosen are reasonable, they don't have a huge effect on the results.

\subsection{Belief Propagation Optimization}

The factor graph defines the dependencies between joint variables and input video frames across time and space in the total probability distribution. With the graph constructed, the next objective is to perform inference to obtain posterior estimates of the 3D joint positions $\{y^t_i\}$.  Belief propagation is a message-passing algorithm which estimates the posterior distribution in CRFs and has been used in many computer vision applications \cite{NIPS2004_2652}.  In the case where the graphs have loops, such as ours, belief propagation is not guaranteed to converge to the true posterior but may be used as an approximation regardless.

As belief propagation will iteratively optimize all frames of a video, an intelligent schedule for organizing the message passing between factors and variables is needed.
Our implementation updates all factor-to-variable messages of the same type before moving on to the next type, i.e. all $f_\mathit{temp}$ messages are updated, then all $f_\mathit{col}$ messages, etc.
The temporal factor update passes messages forward in time through the whole video clip, then back again, so that temporal information is rapidly propagated to distant frames.
Future work could easily adapt this schedule to process live video by temporally optimizing over a sliding window of frames rather than the entire video clip.
We run all of our tests with five iterations of belief propagation across the full network.  We experimentally found that running more than five iterations did not change the values because by the fifth iteration the values had effectively converged.




\section{Experiments}

We evaluate our 3D pose estimation pipeline on the TUM Campus and TUM Shelf \cite{Amin2013} datasets, which have labeled 3D ground truth for multiple people moving in the scene.  The inclusion of multiple people adds an extra dimension of challenge not present in many other 3D pose estimation datasets\cite{h36m_pami, Sigal2009}, because as the people move and interact it is often the case that they obscure each other.

The TUM Campus and Shelf datasets both feature scenes shot from 3 calibrated cameras containing 2-3 people at any given time.  We evaluated the Campus dataset on 220 frames of video with resolution $360 \times 288$ and the Shelf dataset on 300 frames with resolution $1032 \times 776$.  During these test portions of the sequence, actors frequently come in and out of view, occlude each other, and stand stationary for long periods of time.  


To measure the performance of our algorithm, we use the PCP (percentage of correctly estimated parts) metric, following the example set by \cite{Amin2013}.  Using the PCP metric, originally described by \cite{Ferrari2008}, two adjacent joints are considered to be correctly estimated if their respective distances from their ground truth locations are less than $\alpha=0.5$ of the length of the limb in the ground truth. 

\subsection{Comparison to Other Methods}

\begin{table*}                                                   
\centering                                                      
\begin{tabular}{c c c c c c c c|c|c }     
\multicolumn{10}{c}{\textbf{Campus}} \\                                                    
& & Head & Torso & Upper Arm & Forearm & Thigh & Shin & All & Average \\
                                            \hline       
& Actor1 & 64.58 & 100.00 & 94.80 & 66.67 & 100.00 & 81.25 & 85.00 \\   
Amin \etal\cite{Amin2013} & Actor2 & 78.84 & 100.00 & 84.66 & 27.25 & 98.15 & 83.33 & 76.56& 76.61 \\                                 
& Actor3 & 38.52 & 100.00 & 83.71 & 55.19 & 90.00 & 70.37  & 73.70\\ 
\hline
                                                    
& Actor1 &96.55 & 93.10&  96.55&  86.21 & 93.10 & 96.55& \textbf{93.45} \\
Belagiannis \etal\cite{Belagiannis16} & Actor2 & 98.24&  48.82 & 97.35&  42.94 & 75.00&  89.41 & 75.61 & 81.08 \\                                 
& Actor3 & 93.20&  85.44&  89.81 & 74.76&  91.75 & 76.21&  84.37 \\ 

\hline                                                     
& Actor1 & 93.75 & 100.00 & 80.21 & 48.96 & 100.00 & 100.00 & 86.55 \\    
Ours & Actor2 & 47.37 & 98.95 & 91.05 & 42.89 & 98.95 & 98.95 & \textbf{82.54} & \textbf{85.15} \\
& Actor3 & 69.57 & 100.00 & 90.00 & 64.35 & 100.00 & 100.00 & \textbf{88.14} \\  

\multicolumn{10}{c}{\textbf{Shelf}} \\                       
\hline
& Actor1 & 93.75&  100.00 & 73.08&  32.99 & 85.58 & 73.56&  72.42 \\
Amin \etal\cite{Amin2013} & Actor2 & 100.00  &100.00 & 73.53 & 2.94&  97.06&  73.53  &69.41 & 77.3 \\                                 
& Actor3 & 85.23&  100.00 & 86.62&  60.31&  97.89 & 88.73&  85.23\\ 
\hline
                                                      
& Actor1 &96.29 & 100.00 & 82.24 & 66.67 & 43.17 & 86.07 & 75.26 \\
Belagiannis \etal\cite{Belagiannis16} & Actor2 & 78.95 & 100.00 & 82.58&  47.37 & 50.00 & 78.95&  69.68 & 79.00 \\                                 
& Actor3 & 98.00 & 100.00 & 93.15 & 92.30 & 56.50 & 97.00 & 87.59 \\

\hline                                                   
& Actor1 & 32.14 & 99.64 & 95.18 & 82.50 & 99.64 & 99.64& \textbf{88.34} \\
Ours & Actor2 & 16.22 & 100.00 & 94.59 & 66.22 & 100.00 & 100.00 &\textbf{ 85.26} & \textbf{88.92} \\
& Actor3 & 28.40 & 100.00 & 94.14 & 94.44 & 100.00 & 100.00 & \textbf{91.30} \\  
\end{tabular}                                                   
\caption{Performance breakdown by part for on the Campus and Shelf datasets.  Best results are in bold.  The ``Average'' column is weighted by how often each actor appears in the dataset.}                                        
\label{table:campusBreakdown}                                      
\end{table*}

In this section, we compare the performance of our system to other state-of-the-art methods on the Shelf and Campus datasets, as broken down by individual joints.  The results of these experiments are shown in Table \ref{table:campusBreakdown}. 

With the exception of Actor 1 in the Campus dataset, our method out-performs all other methods on the six actors of the Campus and Shelf datasets.  Within the Campus dataset, we find that the method from \cite{Belagiannis16} performs slightly better on each actor's arms, but that our method makes up the difference by performing better on the legs and torso.  We attribute this to differences in the output of the 2D part detectors.

On the other hand, our method performs almost unilaterally better on the Shelf dataset.  This discrepancy is likely due to the resolution difference between the two datasets; at a native resolution of $360 \times 288$, images from the Campus dataset must be scaled up to be used by the OpenPose 2D pose detector, which processes input at a resolution of $656 \times 368$.  The Shelf dataset's $1032 \times 776$ images, on the other hand, can make use of the full discriminative power of OpenPose's part detector.  


\begin{figure}
\centering
\includegraphics[width=.3\linewidth]{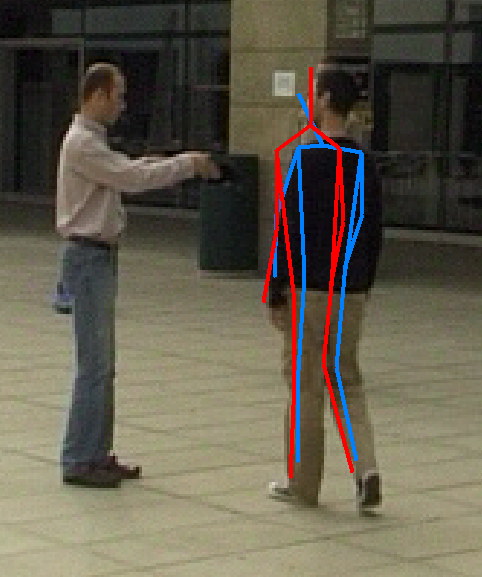}
\includegraphics[width=.15\linewidth]{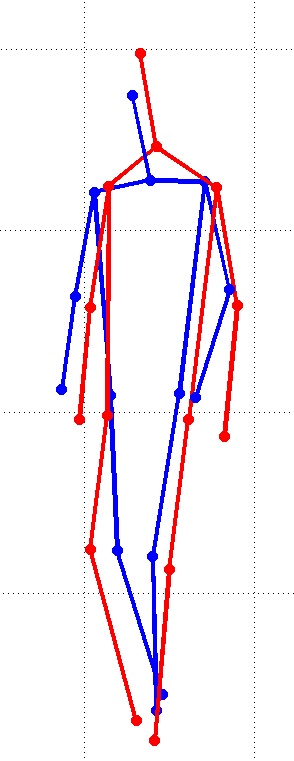}
\caption{A superimposed comparison of our method's output (blue) with the ground truth pose (red) on a single frame of the Campus dataset.  Note that the head and neck joints of our output are systematically lower than on the ground truth, negatively affecting our accuracy. }
\label{fig:skeletonComp}
\end{figure}

During computation of these metrics, the 3D head pose output by our system was manually adjusted by a fixed amount to account for a discrepancy between the OpenPose and ground truth body models.  OpenPose is trained on the Microsoft COCO 2016 Dataset \cite{DBLP:journals/corr/LinMBHPRDZ14} to detect the head joint around the level of the nose, whereas the ground truth labels the head joint as the top of head.  Similarly, OpenPose detects the neck as lying directly between the shoulder blades, as opposed to just above them.  An example is shown in Figure \ref{fig:skeletonComp}. To perform a fair comparison with ground truth, the head and neck detections output by our system were translated upward in the $z$ direction by 10cm for evaluation.  This approximates the correct offset in these datasets since the range of actor pose is limited to upright poses.  In the future we may extend the systems to learn the correct offsets, but longer term we expect that a community standardized human body model will emerge to eliminate these types of discrepancies in future datasets.

We also note that while Belagiannis \etal do out-perform our method on one actor out of six, our method always achieves accuracy of at least 82\%, whereas the other methods we compare against score below 80\% on at least half of the actors, making our method more consistent across sequences.
\begin{figure}
\centering
\begin{subfigure}[t]{.15\textwidth}
\centering
\includegraphics[width=.6\linewidth]{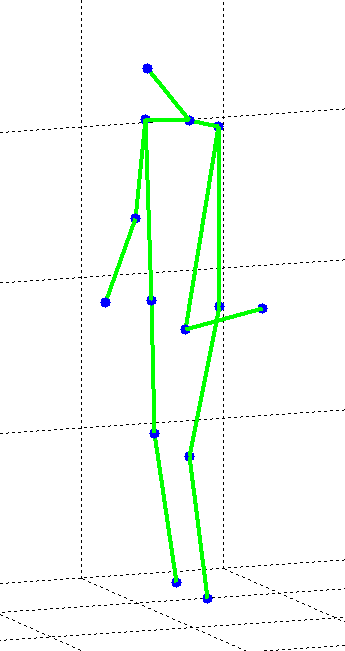}
\caption{Data Only}
\end{subfigure}
\qquad
\begin{subfigure}[t]{.15\textwidth}
\centering
\includegraphics[width=0.5742\linewidth]{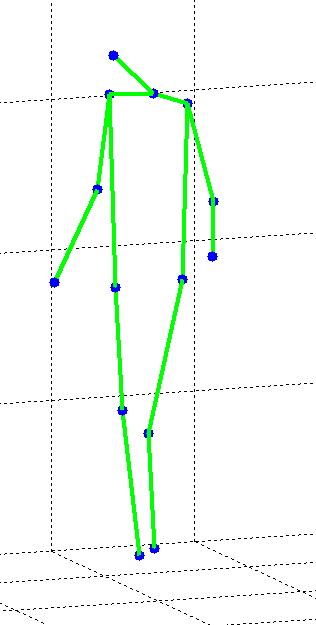}
\caption{All Terms}
\end{subfigure}
\caption{The benefits of temporal smoothing.  (a) Using only the data term, it's possible for certain joints to be incorrectly placed on isolated frames.  (b) The temporal term can propagate information from nearby frames to prevent these mistakes.}
\label{fig:tempSmoothExample}
\end{figure}

\subsection{Analysis of Individual Factors}

Table \ref{table:campusBreakdown} breaks down the effect of each term in the factor graph on our final result accuracy.  When the data term alone is used, belief propagation is not run and the 3D pose hypothesis with the highest average 2D pose likelihood is selected for the answer.  As such, the data-only numbers reflect the performance of a trivial triangulation of the 2D pose output without any additional reasoning.  We note that performance with the data term is competitive with other state-of-the-art methods, and in the case of the Shelf dataset actually out-performs all other methods.  This strong performance can be attributed to the highly effective 2D part detection of OpenPose.

The main contribution of the temporal term on performance is its ability to correct errors that exist only in an  isolated frame.  Figure \ref{fig:tempSmoothExample} illustrates this; errors are corrected by information flow from adjacent frames which do not exhibit the same flaws.  Additionally, when viewing animated videos of the body-models estimated with and without temporal terms side-by-side, we observe that the temporal result is significantly less jittery.  This smoothness is critical to applications such as motion capture, but is not well-captured by the PCP metric.

The collision term has a very minor effect in these results and corrects a small number of minor errors caused by joints being improperly disambiguated.  Its effect may be more pronounced in datasets with more complex human motions not captured in the Campus and Shelf datasets.  However, we conclude that for simple applications it could likely be removed for a modest improvement in runtime at a negligible accuracy penalty.

\begin{table}                                
\centering                                   
\begin{tabular}{c|c c c c}     
\multicolumn{5}{c}{\textbf{Campus}} \\                            
\multicolumn{1}{c}{} & \multicolumn{1}{c}{\textit{Data}} & \multicolumn{1}{c}{\textit{Temp + Data}} & \multicolumn{1}{c}{\textit{Col + Data}} & \multicolumn{1}{c}{\textit{All}} \\
\hline
Actor1 & 76.70 & 86.55 & 76.89 & 86.55 \\               
Actor2\tablefootnote{For one frame of the video, our person detection step failed to identify Actor 2.  For that frame, we supplied a PCP of 0 for all limbs.} & 81.00 & 82.44 & 81.00 & 82.54 \\    
Actor3 & 87.27 & 88.06 & 87.35 & 88.14 \\

\multicolumn{5}{c}{\textbf{Shelf}} \\                          
\hline
Actor1 & 86.75 & 88.31 & 86.66 & 88.34 \\  
Actor2 & 84.52 & 85.26 & 84.77 & 85.26 \\
Actor3 & 90.68 & 91.30 & 90.68 & 91.30 \\
\end{tabular}                        
\caption{The effects of each term on the results for the Campus and Shelf datasets.}                     
\label{table:campusFactors}                   
\end{table}


\section{Conclusion}

We advance the previous state-of-art in multi-person multi-camera pose estimation by combining a modern 2D pose detector based on deep networks with a factor graph optimization to reason over multiple views. We place fewer constraints on limb interactions than previous techniques, yet achieve higher metrics in the Campus and Shelf evaluation datasets. Future work will include adapting the factor graph to allow processing of live video by iterating over short sliding temporal windows.

\section*{Acknowledgements}

The research was supported in part by the Office of Naval Research (N00014-16-P-2040).


%



\bibliography{sources}
\bibliographystyle{IEEEtran}
\end{document}